\pdfoutput=1
\documentclass[11pt]{article}
\usepackage[]{acl} %
\usepackage{adjustbox}
\usepackage{array}
\usepackage{amsmath} 
\usepackage{bm}
\usepackage{booktabs}
\usepackage{color}
\usepackage{enumitem}
\usepackage[utf8]{inputenc}
\usepackage[T1]{fontenc}
\usepackage{graphicx}
\usepackage{latexsym}
\usepackage{listings}
\usepackage{makecell}
\usepackage{microtype}
\usepackage{multirow}
\usepackage{natbib}
\usepackage{soul}
\usepackage{subcaption}
\usepackage{times}
\usepackage{url}
\usepackage{xcolor}
\usepackage{xspace}
\newcommand{\authorspace}{\hspace{0.3cm}}

\newcommand{\approach}{PURR\xspace}
\newcommand{\prevapproach}{RARR\xspace}

\newcommand{\tabitem}{\textbf{-}\xspace}
\definecolor{correct-edit}{HTML}{6AA84F}

\definecolor{wrong-edit}{HTML}{EA4335}

\DeclareMathOperator{\avg}{avg}
\DeclareMathOperator{\maxx}{max}

\newif\ifcomments
\ifcomments
    \newcommand{\ac}[1]{{\color{blue}{\bf{[anthony]}} \emph{#1}}}
\else
    \providecommand{\ac}[2][]{}
\fi

\title{
    \approach{}: Efficiently Editing Language Model Hallucinations \\ by Denoising Language Model Corruptions
}
\author{
    Anthony Chen$^{1}$\thanks{~~Work started during an internship at Google Research.}\authorspace
    Panupong Pasupat$^{2}$\authorspace
    Sameer Singh$^{1}$\authorspace
    Hongrae Lee$^{3}$\authorspace
    Kelvin Guu$^{1}$ \\
    $^{1}$University of California, Irvine \qquad $^{2}$Google DeepMind \qquad $^{3}$Google Search \\
    \small{\{\href{mailto:anthony.chen@uci.edu}{\tt anthony.chen},
      \href{mailto:sameer@uci.edu}{\tt sameer}\}\tt@uci.edu} \qquad
    \small{\{\href{mailto:ppasupat@google.come}{\tt ppasupat},
      \href{mailto:hrlee@google.com}{\tt hrlee},
      \href{mailto:kguu@google.com}{\tt kguu}\}\tt@google.com}
}

\begin{document}
\maketitle
\begin{abstract}
    The remarkable capabilities of large language models have been accompanied by a persistent drawback: the generation of false and unsubstantiated claims commonly known as ``hallucinations''.
    To combat this issue, recent research has introduced approaches that involve editing and attributing the outputs of language models, particularly through prompt-based editing.
    However, the inference cost and speed of using large language models for editing currently bottleneck prompt-based methods.
    These bottlenecks motivate the training of compact editors, which is challenging due to the scarcity of training data for this purpose.
    To overcome these challenges, we exploit the power of large language models to introduce corruptions (\textit{i.e.}, noise) into text and subsequently fine-tune compact editors to denoise the corruptions by incorporating relevant evidence.
    Our methodology is entirely unsupervised and provides us with faux hallucinations for training in any domain.
    Our \textit{Petite Unsupervised Research and Revision} model, \approach{}, not only improves attribution over existing editing methods based on fine-tuning and prompting, but also achieves faster execution times by orders of magnitude.\footnote{The data generation pipeline, training data, and \approach{} checkpoints will be released.}

\end{abstract}
\begin{figure*}[!t]
    \centering
    \begin{subfigure}[b]{0.547\textwidth}
        \centering
        \includegraphics[width=\textwidth]{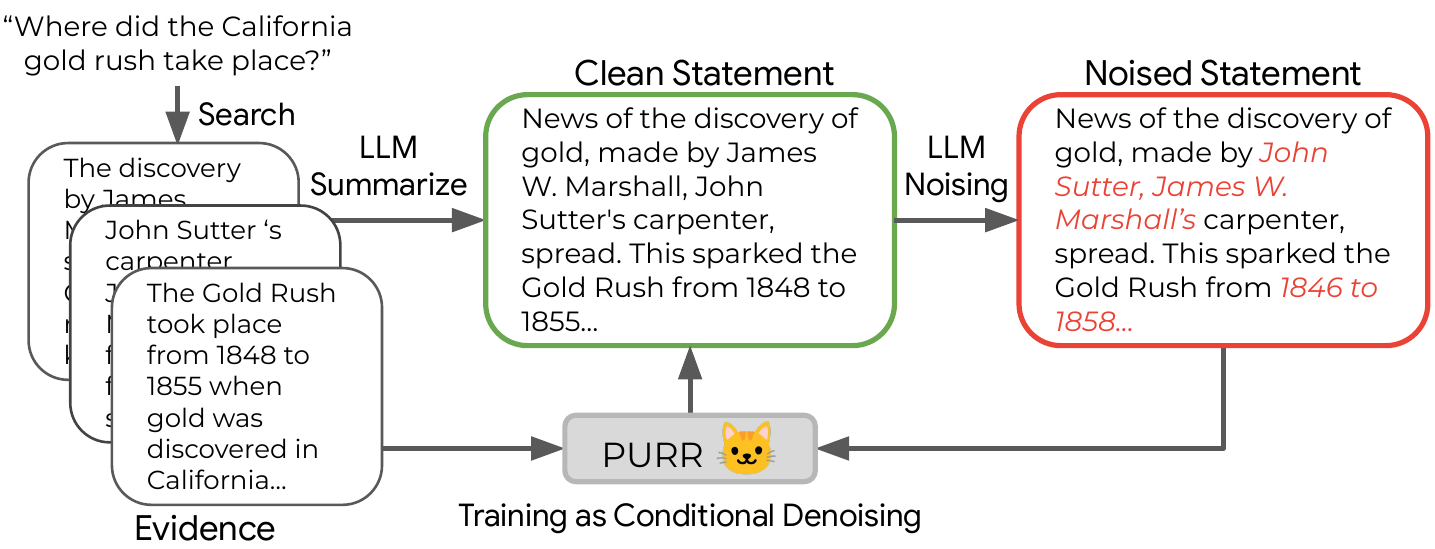}
        \caption{
            \textbf{Training \approach{}.}
            Given a seed query, we search for relevant evidence and summarize them into a claim which we corrupt.
            \approach{} is trained to denoise the corruption conditioned on the evidence.
        }
        \label{fig:training_pipeline}
    \end{subfigure}
    \hfill
    \begin{subfigure}[b]{0.412\textwidth}
        \centering
        \includegraphics[width=\textwidth]{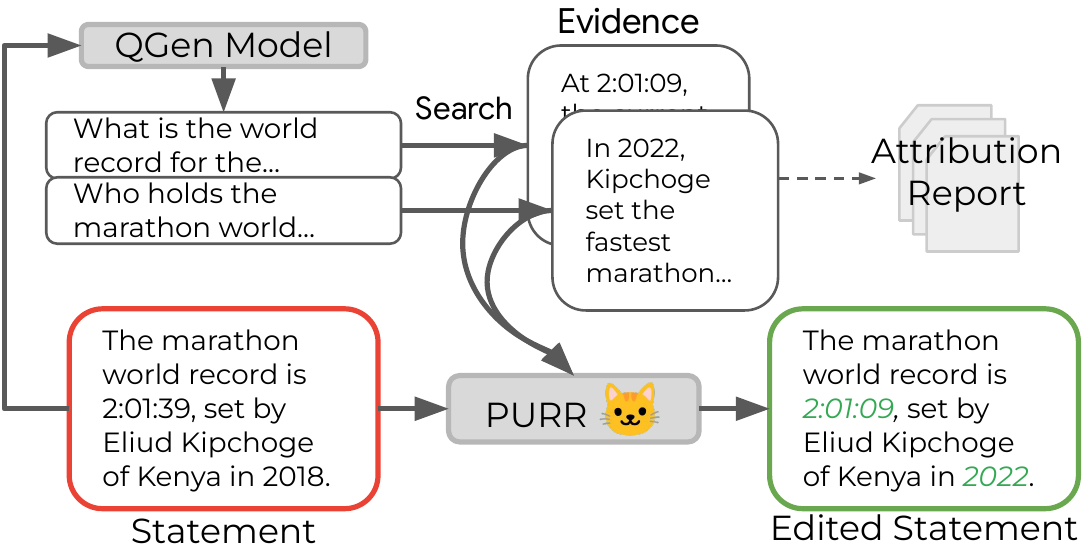}
        \caption{
            \textbf{Using \approach{}.}
            Given an ungrounded statement, we generate questions to search for relevant evidence which is then used to produce an edit.
        }
        \label{fig:inference_pipeline}
    \end{subfigure}
    \caption{\textbf{Training and Using \approach{}.}}
\end{figure*}
\section{Introduction}\label{sec:introduction}

As the strengths of large language models (LLMs) have become prominent \citep{Brown2020LanguageMA, Chowdhery2022PaLMSL, Touvron2023LLaMAOA}, so too have their weaknesses \citep{Bender2021OnTD}.
A glaring weakness of LLMs is their penchant for generating false, biased, or misleading claims in a phenomena broadly referred to as ``hallucinations'' \citep{Maynez2020OnFA,krishna-etal-2021-hurdles, Longpre2021EntityBasedKC,raunak-etal-2021-curious}.
Most LLMs also do not ground their generations to any source, exacerbating this weakness \citep{Rashkin2021MeasuringAI}.

Post-hoc attribution and edit strategies offer promising solutions to tackle the problems of grounding and hallucination in language models \citep{Thorne2020EvidencebasedFE, Gao2022RARRRA}.
These approaches retrieve supporting evidence to attribute the output (referred to as a claim) of a language model, followed by an editor that corrects factual errors in the claim, ensuring consistency with the evidence.
A notable advantage of post-hoc methods is their modularity: they can be easily applied to any text regardless of their generation source.
However, existing editors exhibit distinct strengths and weaknesses.
Sufficiently large language models can be few-shot prompted to perform editing \cite{bai2022constitutional, Gao2022RARRRA}.
However, there is currently a steep compute-quality tradeoff, where only the largest, most expensive models can perform this task well.
Even then, significant quality headroom remains, as we will show.
In contrast, much smaller, cheaper models can be fine-tuned to perform editing, but are limited to specific domains where adequate training data is available \citep{iv-etal-2022-fruit, Schick2022PEERAC}.

Instead of utilizing LLMs as prompted editors, we leverage their general-purpose capabilities to introduce challenging corruptions (\textit{i.e.}, noise) to clean pieces of text.
Subsequently, we fine-tune compact editors to denoise these corruptions by grounding onto relevant evidence.
While text to corrupt is readily available, we do not assume that paired relevant evidence is provided.
To tackle this, our data generation pipeline first searches for a collection of topically related evidence.
We then employ an LLM summarize the evidence into a claim which is then noised (Fig. \ref{fig:training_pipeline}).
The evidence is then used to ground the denoising.
In contrast to existing work that assumes access to relevant paired evidence to ground the edit when training \citep{Balachandran2022CorrectingDF} or assumes edit data is provided for training \citep{Schick2022PEERAC, iv-etal-2022-fruit}, our approach eliminates these assumptions. 
Furthermore, unlike distillation where a challenging distillation set is vital and the student model generally under-performs the teacher \citep{Beyer2021KnowledgeDA, Stanton2021DoesKD}, our noising process introduces challenging corruptions and our resulting editor trained on these corruptions surpasses the performance of the LLM used for noising when the same LLM is employed for prompted editing on multiple datasets.

Our \textit{Petite Unsupervised Research and Revision} model, \approach{}, is built by fine-tuning a fusion-in-decoder T5 model on denoising data from our data generation pipeline \citep{raffel_exploring_2020, izacard-grave-2021-leveraging}.
Because our goal is to improve attribution broadly across tasks and domains, we evaluate \approach{} on outputs of large language models on multiple question answering and dialog datasets.
On all benchmarks, \approach{} outperforms much more expensive LLM-prompted editors in improving attribution while being orders of magnitude faster.

\section{Editing for Attribution}\label{sec:task}
\subsection{Task Overview}
While there are various ways to apply editing to the outputs of large language models, the primary objective of \approach{} is to present efficient methods for attributing the outputs of language models and rectifying inaccuracies, referred to as \textit{Editing for Attribution} \citep{Gao2022RARRRA}.
In this task, a system is provided with a textual statement, $x$, and is tasked to produce an \textit{attribution report}.
The attribution report consists of a collection of evidence snippets, $A = \{e_1, e_2, \ldots, e_n\}$, that grounds the information in $x$.
Additionally, the system is asked to produced a revised statement (\textit{i.e.}, edit), $y$, that fixes any inaccuracies in $x$ that contradict the content in $A$.
For completeness, we present a summary of the task and refer interested readers to \citet{Gao2022RARRRA} for a more comprehensive discussion.

\subsection{Evaluation Metrics}
Following \citet{Gao2022RARRRA}, we evaluate editing-for-attribution systems along two dimensions: \textbf{attribution}, the extent to which the original and revised statements can be attributed to the attribution report, and \textbf{preservation}, which measures how much information has changed from $x$ to $y$.
The objective of the task is to maximally attribute a textual statement while preserving the original intent of the language model generation to the greatest extent possible.
We use automated metrics developed by \citet{Gao2022RARRRA} to measure both attribution and preservation, which were found to have strong correlation to human raters.
It is important to note that this evaluation setup does not require reference edits and only relies on the grounding between the textual statements and the attribution report.

\paragraph{Attribution}
A textual statement is generally said to be attributable to a set of evidence if one could reasonably say that given the evidence set, the statement is entailed \citep{Rashkin2021MeasuringAI}.
To formalize this, \citet{Gao2022RARRRA} introduce an evaluation metric based on sentence-level natural langauge inference (NLI) model.
Given an attribution report, $A$, and a textual statement $y$ consisting of sentences, $y = \{s_1, s_2, \ldots\}$, we use a NLI model to measure the likely that each sentence is entailed by an evidence snippet in $A$: $\text{NLI}(e, s_i)$.
The attribution of the entire statement, $y$, is computed as the average over the maximum attribution score for each constituent sentence.
\begin{equation}
\text{Attr}_{(s, A)} = \maxx_{e \in A} \text{NLI}(e, s)
\end{equation}
\vspace{-5mm}
\begin{equation}
\text{Attr}_{(y, A)} = \avg_{s \in y} \text{Attr}_{(s, A)}
\end{equation}
The goal of editing is to have $\text{Attr}_{(y, A)}$ be higher than $\text{Attr}_{(x, A)}$.

\paragraph{Preservation}
Preservation is measured using character-level Levenshtein distance between $x$ and $y$.
Preservation is 1 if the statements are the same and 0 if $y$ has completely changed all textual information in $x$.
\begin{equation}
\text{Pres}_{(x, y)} = \max\left(1 - \frac{\text{{Lev}}(x, y)}{\text{{length}}(x)}, 0\right)
\end{equation}

To capture our goal of maximal attribution with maximal preservation, we unify these two metrics by computing the harmonic mean, $F1_{AP}$, of $\text{Attr}_{(y, A)}$ and $\text{Pres}_{(x, y)}$.

\subsection{Evaluation Sets}
Our goal is to improve attribution broadly across tasks and domains on the outputs of strong generations systems.
\citet{Gao2022RARRRA} construct evaluation sets by prompting strong LLMs to generate outputs on three tasks: Natural Questions (factoid question answering) \citep{kwiatkowski-etal-2019-natural}, StrategyQA (reasoning-chain question answering) \citep{geva-etal-2021-aristotle}, and QreCC (knowledge-intensive dialogue) \citep{anantha-etal-2021-open}.
\citet{Gao2022RARRRA} generate 150 validation and 150 test instances for each dataset using PALM for Natural Questions and StrategyQA and LaMBDA on QReCC \citep{Chowdhery2022PaLMSL, Thoppilan2022LaMDALM}.
We use these sets and tune on the validation sets and report results on the test sets.

\subsection{Baselines} 

\approach{} and all baselines follow a \textbf{research-and-revision} pipeline.
In the \textbf{research} stage, the objective is to search for relevant pieces of evidence to ground the information in the textual statement, $x$.
This stage remains consistent across all baselines.
We first prompt a large language model to generate a set of queries $Q = \{q_1, q_2, \ldots\, q_m\}$ that attempts to cover all pieces of information in $x$ that needs verification.
Subsequently, we use Google Search in conjunction with a passage extractor to find the most relevant evidence snippet for each query, constituting an evidence set $E = \{e_1, e_2 \ldots, e_m\}$.

In the \textbf{revision} stage, an editor is given the original statement, $x$, the set of queries, $Q$, and the evidence set, $E$, and asked to produce a revised statement, $y$.
$y$ can be the same as $x$ in the event the editor deems the original statement cannot be edited further to increase attribution.
We measure the ability of different editors to abstain from editing later on.
We compare \approach{} against two baseline editors.

\paragraph{EFEC} is a fine-tuned T5 editor trained on FEVER \citep{fever-2021-fact}.
EFEC was trained using evidence retrieved from Wikipedia and concatenates all pieces of evidence with the text statement to produce an edited statement.
Notably, EFEC does not use the query set when making an edit.
\cite{Gao2022RARRRA} found EFEC often improves attribution at the expense of preservation.

\paragraph{\prevapproach{}} is a prompt-based editing approach that builds upon PALM, a language model with 540 billion parameters \citep{Chowdhery2022PaLMSL}.
Unlike EFEC, which incorporates all evidence simultaneously to produce an edit, \prevapproach{} iteratively examines each evidence, $e_i$, by checking whether there is any contradictions between the text statement, $x$, and edits in the event there is.
The process of contradiction checking and editing is performed using distinct few-shot prompts.
\citet{Gao2022RARRRA} demonstrate that this iterative approach to editing combined with few-shot prompting leads to improvements in attribution and preservation, albeit at the cost of multiple computationally expensive and slow calls to a large language model.

\subsection{Generating the Attribution Report}
To maintain a manageable scope, we limit the attribution report, $A$, to include only the five most relevant pieces of evidence from the evidence set, $E$.
An attribution report of five evidence snippets was found to be able to attribute the information for the claims in the datasets we evaluate on.
It is worth noting that when editing, there are no restrictions on the number of evidence snippets an editor can utilize.
Given the evidence set, $E$, and the query set, $Q$, from the research stage, we employ a scoring module that evaluates the relevance of each evidence $e_i$ to each query $q_j$, $S(q_i, e_j)$.
Our objective is to identify a subset of evidence that maximizes the coverage across all queries to form the attribution report.
This coverage is quantified as the sum of the highest relevance scores achieved by each query with respect to any evidence.
For scoring, we use a cross-encoder\footnote{\href{https://huggingface.co/cross-encoder/ms-marco-MiniLM-L-6-v2}{https://huggingface.co/cross-encoder/ms-marco-MiniLM-L-6-v2}}.

\begin{equation}
    \text{Cov}_{(E, Q)} = \sum_{i=1}^N \max_{e_j \in E} S(q_i, e_j)
\end{equation}

\section{Efficient Editing by Denoising}\label{sec:method}
In this section, we present an overview of \approach{}, highlight its distinguishing features compared to baselines, and describe the denoising training strategy.

\subsection{Overview of \approach{} at Inference Time}
We first describe how \approach{} is used at inference time and highlight the differences between \approach{} and baselines (Fig. \ref{fig:inference_pipeline}).
Similar to EFEC, \approach{} is built upon on the T5 model, specifically T5-large.
Furthermore, our editing framework adopts a similar approach to EFEC in terms of incorporating all available evidence simultaneously when making an edit.
However, instead of concatenating the evidence in the input, we employ fusion-in-decoder (FiD) to effectively aggregate information across evidence \citep{izacard-grave-2021-leveraging}.
This approach has demonstrated superior performance in merging information and allows us to surpass the context length limits imposed by modern language models.
Finally, rather than employing a prompted language model for query generation during the research stage, we employ distillation to train a T5-large query generation model.
Although our primary focus lies in enhancing the editing process, we opt for distillation during query generation as well to ensure that our editing pipeline does not rely on prompting.

\subsection{Creating Training Data via Noising}
To train an editor to fix hallucinations, we need a dataset consisting of a clean statements, $y$, which are paired with a set of supporting evidence $E=\{e_1, e_2, \ldots, e_n\}$, as well as a corrupted statement, $x$.
While collecting this data manually is feasible, doing so can be expensive, requiring scouring for evidence to ground an LLM generation followed by removing any inaccuracies in the generation.
Instead, we remove this bottleneck by leveraging the general purpose generation capabilities of LLMs to create a training set in a completely fashion.
We generate clean statements by providing a set of topically related evidence to the LLM, and then corrupt the statements to create simulated hallucinations (Fig. \ref{fig:training_pipeline}). 
We provide the prompts used for summarization and corruption in Appendix \ref{sec:appendix:prompts}.

\begin{table}[t!]
    \centering
    \small
    \begin{tabular}{@{\hspace{-10pt}}r@{\hspace{4pt}} p{0.955 \columnwidth}}
        \toprule
        $q$: & Who will be the new coach of the Detroit lions? \\
        $E^+$: & \tabitem On Jan. 20, 2021 the Detroit Lions named Dan Campbell the franchise's new head coach\ldots \\
        & \tabitem Campbell possesses 23 years of NFL experience, including 12 years as a coach and 11 as a player. In his first year\ldots \\
        & \tabitem On Jan. 20, 2021 the Detroit Lions named Dan Campbell the franchise's new head coach\ldots \\
        \textcolor{red}{$\bm{x}$}/$y$: & Dan Campbell was appointed the new \st{head} \textbf{\textit{\textcolor{red}{assistant}}} coach of the Detroit Lions on January 20, 2021. With \st{23} \textbf{\textit{\textcolor{red}{19}}} years of NFL experience, 12 as a coach and \st{11} \textbf{\textit{\textcolor{red}{7}}} as a player\ldots \\
        \midrule
        $q$: & What is the neurological explanation for why people laugh when they're nervous or frightened? \\
        $E^+$: & \tabitem A 2015 Yale study found people respond with a variety of emotions to strong outside stimuli\ldots \\
        & \tabitem Vilayanur Ramachandran states ``We have nervous laughter because we want to make ourselves think what horrible thing we encountered isn't really as horrible as it appears''\ldots\\
        & \tabitem Stanley Milgram conducted one of the earliest studies about nervous laughter in the 1960s. His study revealed that people often laughed nervously in uncomfortable situations\ldots\\
        \textcolor{red}{$\bm{x}$}/$y$: & Yale researchers in 2015 found people often respond to strong external stimuli with a variety of emotions, including \st{nervous laughter} \textbf{\textit{\textcolor{red}{anger}}}. \st{Stanley Milgram's} \textbf{\textit{\textcolor{red}{Vilayanur Ramachandran's}}} 1960s study also observed this in uncomfortable situations. Neuroscientist \st{Vilayanur Ramachandran} \textbf{\textit{\textcolor{red}{Stanley Milgram}}} theorizes that people laugh when\ldots. \\
        \bottomrule
    \end{tabular}
    \caption{
        \textbf{Training Examples}.
        Our editing data covers a variety of domains and introduces challenging corruptions (\textit{e.g.,} numerical, entity, and semantic role).
        $q$ is the seed query, $E^+$ is the gold evidence set used to generate the clean statement, $y$ is the clean statement and \textit{\textbf{\textcolor{red}{\bm{$x$} is the corrupt statement}}}.
        \label{tab:examples}
    }
\end{table}

\paragraph{Generating Clean Statements With Evidence}
The first step is to create a statement, $y$, paired with a set of evidence, $E$, that attributes (\textit{i.e.,} grounds) the statement.
Our pipeline only requires a set of queries in the domain of interest to get started.
We start with a query, $q$, and use a search engine to find evidence related to the question.
We take the top web pages from the search engine and chunk them into passages.
Using the same cross-encoder from the attribution report scoring module, we bin the passages that have the highest relevant score (beyond some threshold) to $q$ into a set of gold evidence $E^+ = \{e^+_1, e^+_2, \ldots, e^+_i\}$ and the rest of the passages into a set of hard negative evidences $E^- = \{e^-_1, e^-_2, \ldots, e^-_j\}$.
In our pipeline, we restrict the size of $E^+$ to contain at most four pieces of evidence.
The resulting evidence set is the union of the gold and hard negative evidences $E = E^+ \cup E^-$.
We then prompt a large language model to do zero-shot multi-document summarization of the gold evidence set, $E^+$.
We use the resulting summary as the clean statement, $y$, and upon manual inspection, the summary has a high degree of faithfulness to the evidence set.
 
\paragraph{Noising and Conditional Denoising}
We take the clean statement, $y$, and noise it by prompting a large language model to corrupt the text resulting in the corruption $x$.
Our prompt contains examples of corruptions, and covers a wide range of linguistic phenomena we observe when it comes to LLM hallucinations.
These include incorrect dates and entities, semantic role errors, and quantification errors.
Once noised claims paired with evidence is available, an editor can be trained by fine-tuning a sequence-to-sequence model to maximize $P(y | x, E)$.
We call the resulting editor from denoising \approach{}.

\subsection{Dataset Statistics and Training Details}
We utilized GPT-3.5 \texttt{text-davinci-003} to facilitate the process of generating summaries and introducing corruption.
Our choice of this particular model ensures that our generation strategy can be easily replicated.
We started with roughly 6,000 seed queries covering a variety of domains and topics resulting in an edit dataset of 6,000 instances (Tab. \ref{tab:examples}).
We reserve 10\% for validation and use the resulting 90\% for training.
Each instance cost roughly 4 cents to generate and in total cost of roughly \$250.

We fine-tune T5-large on our dataset using the validation loss to tune hyperparameters and determine training stoppage.
During training, we pair each corrupted statement, $x$, with four pieces of evidence from the accompanying gold evidence set, $E^+$, to ground the edit and produce the clean statement, $y$.
In the event that the gold evidence set has fewer than four evidence snippets, we randomly sample evidence from the negative evidence set, $E^-$, until we hit four snippets.
We found adding negative evidence during training helps \approach{} ignore irrelevant evidence during inference.

\section{Results}\label{sec:results}

\subsection{Primary Quantitative Results}
We provide results on the editing-for-attribution task in Table \ref{tab:attribution_results}.
We report the attribution of the claim before and after editing and the preservation of the edited claim.
Our primary metric, $F1_{AP}$, is the harmonic mean between the attribution and preservation of the edited claim.
We first turn our attention to the baselines.
EFEC, the editor that was fine-tuned with evidence largely from Wikipedia, struggles on this task.
While EFEC improves attribution, this comes at the large expense of preservation and we see this in practice as EFEC tends to make large changes to the claim.
\prevapproach{}, the prompted editor, does not improve attribution as much as EFEC.
However it is significantly better at preserving the intent of the original claim.
Because of this, \prevapproach{} is much better on the unified $F1_{AP}$ metric.

\approach{} improves upon the results of \prevapproach{} by generally making smaller changes to the claim while improving the attribution in this more limited edit.
Because of this, \approach{} pushes the state-of-the-art on the unified $F1_{AP}$ metric an all three tasks.
Moreover, \approach{} is significantly more efficient to use by virtue of its size.

\begin{table}[t!]
    \centering
    \adjustbox{max width=\textwidth}{
    \begin{tabular}{lcc | c}
    	\toprule
    	Model & Attr. ($x \rightarrow y$) & Pres. & $F1_{AP}$ \\
    	\midrule
    	& \multicolumn{3}{c}{\bf PALM outputs on NQ}\\
        EFEC & 44.7 $\rightarrow$ \textbf{63.9} & 39.6 & 48.5 \\
        \prevapproach{} & 44.7 $\rightarrow$ 53.8 & 89.6 & 67.2 \\
    	\approach{} & 44.8 $\rightarrow$ 59.8 & \textbf{91.0} & \textbf{72.2} \\
    	\midrule
    	& \multicolumn{3}{c}{\bf PALM outputs on SQA} \\
    	EFEC & 37.2 $\rightarrow$ \textbf{58.2} & 31.0 & 40.4 \\
        \prevapproach{} & 37.2 $\rightarrow$ 44.6 & 89.9 & 59.6 \\
    	\approach{} & 36.9 $\rightarrow$ 47.1 & \textbf{92.0} & \textbf{62.3} \\
    	\midrule
    	& \multicolumn{3}{c}{\bf LaMBDA outputs on QreCC} \\
    	EFEC & 18.4 $\rightarrow$ \textbf{47.2} & 39.0 & 42.7 \\
        \prevapproach{} & 18.4 $\rightarrow$ 28.7 & 80.1 & 42.2 \\
    	\approach{} & 16.8 $\rightarrow$ 33.0 & \textbf{85.8} & \textbf{47.7} \\
        \bottomrule
    \end{tabular}
    }
    \caption{\textbf{Results on the \textit{Editing for Attribution} task.}
        We report the attribution of the statement before and after editing, preservation after editing, and $F1_{AP}$ which combines attribution and preservation.
        Results are on LLM outputs on factoid question answering, long reasoning question answering, and dialog.
    }
    \label{tab:attribution_results}
\end{table}

\subsection{Breaking Down the Numbers}
We dig into the edits to get a better sense of where \approach{} improves on the baselines.
Based on the preservation, $\text{Pres}_{(x, y)}$, and attribution scores of the original statement, $\text{Attr}_{(x, A)}$, and edited statement, $\text{Attr}_{(y, A)}$, we say an edit can fall into one of the following sets:
\begin{itemize}[leftmargin=10pt,topsep=2mm,itemsep=0mm]
    \item \textbf{Huge Edit}: We say an edit is ''huge`` if preservation is low: $\text{Pres}_{(x, y)} < 0.5$.
    \item \textbf{Bad Edit}: We say an edit is ''bad`` if the attribution after editing is lower than before: $\text{Attr}_{(y, A)} - \text{Attr}_{(x, A)} < -0.1$.
    \item \textbf{Unnecessary Edit}: We say an edit is ``unnecessary'' if it is a bad edit and also $\text{Attr}_{(x, A)} > 0.9$. This means the editor made a poor edit when the attribution was already near perfect before editing.
    \item \textbf{Good Edit}: We say an edit is ``good'' if attribution has significantly improved while preservation is high: $\text{Attr}_{(y, A)} - \text{Attr}_{(x, A)} > 0.3$ and $\text{Pres}_{(x, y)} > 0.7$.
\end{itemize}

\begin{figure}[!t]
\centering
\includegraphics[width=\columnwidth]{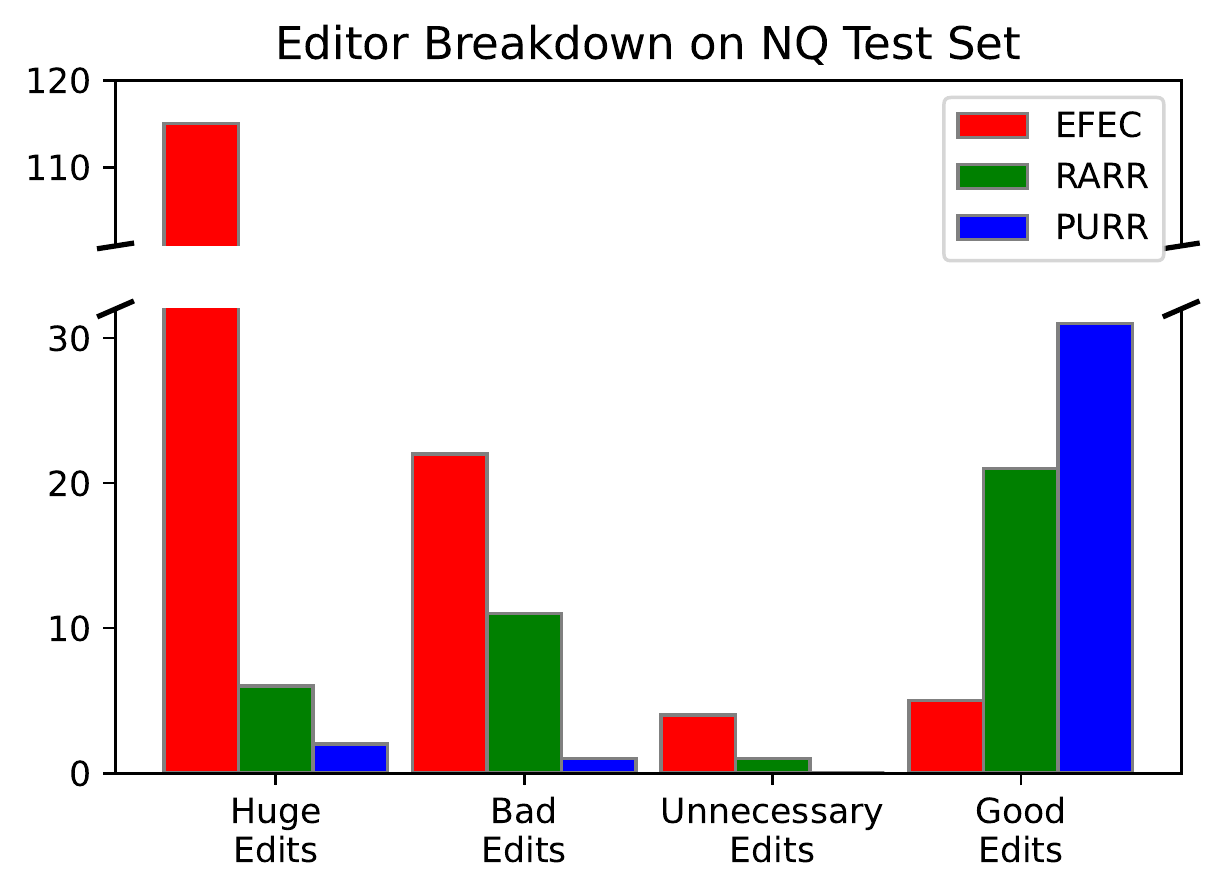}
\vspace{-5mm}
\caption{
\label{fig:edit_counts}
    \textbf{Breakdown of edit types each editor makes} on the Natural Questions test set.
    EFEC makes huge edits while \prevapproach{} sometimes over edits.
    PURR does a much better job at balancing attribution and preservation while rarely over-editing.
}
\end{figure}

Note that unnecessary edits are a subset of bad edits.
We take the 150 instances in the Natural Questions test set and categorize the edits each editor makes in Figure \ref{fig:edit_counts}.
On a majority of claims, EFEC makes large edits while rarely making edits that improve attribution while preserving the original claim.
\prevapproach{} does a much better job at minimizing large edits but there are still cases where \prevapproach{} edits a claim in a way that reduces the attribution.
\approach{} almost never makes large edits and never edits a claim when it is near-perfect in a way that reduces attribution.
\approach{} also makes more good edits compared to the baselines.

\begin{table}[t!]
    \centering
    \small
    \begin{tabular}{@{\hspace{-10pt}}r@{\hspace{4pt}}p{0.957\columnwidth}}
    \toprule
    \multicolumn{2}{c}{\textbf{Correctly revising an entity/span.}} \\
    $x$: & The US stopped printing \$2 bills in 1966. The reason was that vending machines would not accept it. In 1976, the \$2 bill was reintroduced\ldots \\
    $E$: & During that time, most goods and services were less than a dollar, making paper currency impractical. As need and use declined, the Federal Reserve stopped printing \$2 bills in 1966. In 1976, the Federal Reserve brought back the \$2-dollar bill. \\
    $y$: & The US stopped printing \$2 bills in 1966. The reason was that \textcolor{correct-edit}{\textbf{most goods and services were less than a dollar.}} In 1976, the \$2 bill was reintroduced\ldots \\
    & \rule{0pt}{3ex} $Pres_{(x,y)}$: 0.81 \hspace{10pt} $Attr_{(A, x)} \rightarrow Attr_{(A, y)}$: 0.61 $\rightarrow$ 0.84 \\
    
    \midrule
    
    \multicolumn{2}{c}{\textbf{Reasoning across multiple pieces of evidence}} \\
    $x$: &  In 1899, the first driver's license was issued in France. The first license in the United States was issued in 1902 in Massachusetts.\\
    $E$: & \tabitem France introduced the first driving licences under the Paris Police Ordinance of 14 August 1893\ldots\\
    & \tabitem Massachusetts and Missouri were the first states to require a license in 1903, but there was no test associated with it\ldots \\
    $y$: & In \textcolor{correct-edit}{\textbf{1893}}, the first driver's license was issued in France. The first license in the United States was issued in \textcolor{correct-edit}{\textbf{1903}} in Massachusetts. \\
    & \rule{0pt}{3ex} $Pres_{(x,y)}$: 0.90 \hspace{10pt} $Attr_{(A, x)} \rightarrow Attr_{(A, y)}$: 0.03 $\rightarrow$ 0.96 \\
    
    \midrule
    
    \multicolumn{2}{c}{\textbf{Challenging distractors in evidence}} \\
    $x$: &  The first orca put in captivity was in 1965. It was at Seattle, Washington's Seattle Aquarium. The orca's name was Wanda. \\
    $E$: & \tabitem On Nov. 18, 1961 a killer whale had been spotted and a collection crew from the \textbf{Marineland of the Pacific} scrambled to capture her. Never had an orca been captured alive, and Wanda was trying to avoid being the first\ldots\\
    & \tabitem The first orca to be captured in 1961 was found off the coast of California and given the name Wanda. \\
    $y$: & The first orca put in captivity was in \textcolor{correct-edit}{\textbf{1961}}. It was at \textcolor{wrong-edit}{\textbf{Marineland of the Pacific}}. The orca's name was Wanda. \\
    & \rule{0pt}{3ex} $Pres_{(x,y)}$: 0.77 \hspace{10pt} $Attr_{(A, x)} \rightarrow Attr_{(A, y)}$: 0.33 $\rightarrow$ 0.77 \\
    
    \bottomrule
    \end{tabular}
    \caption{
        \ac{add a cooler example here}
        \textbf{Example of \textcolor{correct-edit}{good} and \textcolor{wrong-edit}{bad} revisions with PURR.}
        $x$ = claim; $E$ = relevant evidence; $y$ = edited claim using $E$.
        \approach{} can handle hallucinated entities and spans as well as merge information across evidence to edit.
        \approach{} can struggle when there are challenging distractors in a piece of evidence.
    }
    \label{tab:example-revisions}
\end{table}

\subsection{Qualitative Analysis}
We then dig into the \approach{} predictions and diagnose the strengths of \approach{} and examine where there is room for improvement.
We show examples in Table \ref{tab:example-revisions} that we found are representative of the strengths of \approach{} and areas of potential improvement.
We find that \approach{} is extremely strong at fixing entity and numerical hallucinations as well as longer spans.
Additionally, because \approach{} uses fusion-in-decoder, it is adept at merging information across multiple pieces of evidence to make an edit.
We noticed several instances where there are challenging distractors in evidence that can lead to an erroneous edit.
Future work will introduce stronger corruptions in the data generation pipeline to better handle this case.

We next analyze the entire inference pipeline of \approach{} (Fig. \ref{fig:inference_pipeline}), which includes the question generation model, the search engine, and the editor itself.
Our goal is to see when there is an error, which component is responsible.
On the Natural Questions subset of the evaluation, we examine 20 instances where the attribution after editing, $\text{Attr}_{(y, A)}$, is less than 0.30.
Our qualitative analysis is provided in Table \ref{tab:error-breakdown}.
Roughly 80\% of the instances have low attribution after editing because either the question generation model we used did not fully cover the information in the claim or our search procedure did not find the best evidence for editing.
We believe the question generation is the easier problem to fix while search is a much harder problem.
Editing is a fairly small issue in comparison.
Finally, there are some claims that fall into a ``miscellaenous'' category, either because it was not contextualized enough to properly edit or because the automatic metric erroneously assigned a low score.

\subsection{Inference Speed and Cost Comparisons of Fine-tuned vs Prompted Editors}
A key advantage of \approach{} over prompt-based editors are the lower computational costs.
\prevapproach{}, a prompt-based editor built upon 540B PALM, runs on dozens of TPUs and takes approximately ~40 seconds to edit a single statement.
In comparison, \approach{} can run on a 12GB GPU and takes approximately ~2 seconds to edit a single statement on a Titan-RTX.
Considering generating our training set costs <\$300 USD which is quickly amortized, we recommend our synthetic data generation strategy for large-scale deployment given the speed and cost savings of \approach{}.

\begin{table}[t!]
    \centering
    \small
    \begin{tabular}{@{\hspace{-10pt}}r@{\hspace{4pt}}p{0.957\columnwidth}}
    \toprule
    
    \multicolumn{2}{c}{\textbf{Query Generation Missing Coverage} (\textit{35\%})} \\
    $x$: &  Legends of Tomorrow season 3 finale aired on April 9, 2018. It's title is No Country for Old Dads and is a 42-minute episode.\\
    $Q$: & \tabitem When did the season 3 finale of Legends of Tomorrow air?\\
    & \tabitem \st{\textit{What's the title of Legends of Tomorrow season 3 finale?}} \\
    & \tabitem \st{\textit{How long is the season 3 finale of Legends of Tomorrow?}} \\
    $y$: & Legends of Tomorrow season 3 finale aired on October 10, 2017. It's title is Freakshow and is a 43-minute episode. \\
    & \rule{0pt}{3ex} \hspace{35pt} $Attr_{(A, x)} \rightarrow Attr_{(A, y)}$: 0.01 $\rightarrow$ 0.01 \\
    \midrule
    
    \multicolumn{2}{c}{\textbf{Search Not Finding Best Evidence} (\textit{45\%})} \\
    $x$: & Bryce Dallas Howard plays the character of the mother of Cindy Lou Who in the 2018 animated movie The Grinch. \\
    $E$: &  \st{\textit{Donna Lou Who is one of the characters of the 2018 Illumination animated film, Dr. Seuss' The Grinch. She's the mother of Cindy Lou Who and twin infants sons, Buster and Bean. She's voiced by Rashida Jones .}} \\
    $y$: &  No edit. \\
    & \rule{0pt}{3ex} \hspace{35pt} $Attr_{(A, x)} \rightarrow Attr_{(A, y)}$: 0.05 $\rightarrow$ 0.05\\    
    \midrule
    
    \multicolumn{2}{c}{\textbf{Editor Error/Confusion} (\textit{5\%})} \\
    $x$: &  James Marion Sims was the first to perform the first c section in 1846. This was in the US.\\
    $E$: & \tabitem The first recorded cesarean where both mother and child survived was done in Switzerland, in 1500. That was also a husband-wife affair, although in this case Jacob Nufer was a swine gelder, not a doctor. \\
    & \tabitem The first documented cesarean section on a living woman was performed in 1610; she died 25 days after the surgery. \\
    $y$: & James Marion Sims was the first to perform the first c section in \textcolor{wrong-edit}{\textbf{1610}}. This was in the UK. \\
    & \rule{0pt}{3ex} \hspace{35pt} $Attr_{(A, x)} \rightarrow Attr_{(A, y)}$: 0.00 $\rightarrow$ 0.20 \\    
    \midrule
    
    \multicolumn{2}{c}{\textbf{Miscellaneous} (\textit{15\%})} \\
    $x$: &  The song was written by Tony Macaulay and John Macleod and recorded by Helen Shapiro in 1961.\\
    $y$: & The song was written by Dub Allbritten and Ronnie Self and recorded by The Jackson 5 in 1960.\\[5mm]
    \multicolumn{2}{c}{\textit{Claim not contextualized enough to properly edit.}}\\
    & \rule{0pt}{3ex} \hspace{35pt} $Attr_{(A, x)} \rightarrow Attr_{(A, y)}$: 0.00 $\rightarrow$ 0.01 \\
    \bottomrule
    \end{tabular}
    \caption{
        \textbf{Error Analysis of \approach{} Inference Pipeline.} We sample \textbf{20} edits from the NQ set where attribution is low after editing and categorize why by component.
        $x$ = claim; $Q$ = generated queries used for search, $E$ = relevant evidence; $y$ = edited claim using $E$.
        \st{\textit{Strike-through text}} represents a query that wasn't generated or evidence that wasn't retrieved but should have been.
    }
    \label{tab:error-breakdown}
\end{table}

\section{Related Work}\label{sec:related}

\paragraph{Editing for Attribution}
\approach{} builds upon previous research on post-hoc editing methods aimed at enhancing the attribution and accuracy of generated text \citep{balachandran-etal-2021-simple, cao-etal-2020-factual, iso-etal-2020-fact}.
Notably, RARR \citep{Gao2022RARRRA} and Rethinking-with-Retrieval \citep{he_rethinking_2022} employ few-shot prompting to rectify language model outputs, exhibiting similarities to our work.
FRUIT  \citep{iv-etal-2022-fruit} and EFEC \citep{Thorne2020EvidencebasedFE} also utilize fine-tuned editors to achieve similar objectives, leveraging Wikipedia as a source of training data.
PEER is trained on Wikipedia edits \citep{Schick2022PEERAC} and includes a component for enhancing factuality through editing, but its primary focus lies in collaborative writing. 
Our denoising approach combines the speed advantages of fine-tuned editors while circumventing the reliance on training data that is typically constrained to specific domains like Wikipedia.

\paragraph{Improving Trust in Large Language Models}
Ensuring the safe deployment of large language models encompasses various considerations, beyond just factuality and attribution.
Large language models have demonstrated the potential to regurgitate protected information \citep{Carlini2020ExtractingTD}, spew hateful content \citep{gehman-etal-2020-realtoxicityprompts}, and exhibit high sensitivity to input variations \citep{Zhao2021CalibrateBU}.
A common approach to addressing these issues has been via additional training such as instruction fine-tuning \cite{Sanh2021MultitaskPT, Min2021MetaICLLT, Chung2022ScalingIL, Ye2022GuessTI}, fine-tuning from human feedback \citep{Ziegler2019FineTuningLM, Stiennon2020LearningTS}, and more recently pre-training from human feedback \citep{Korbak2023PretrainingLM}.
In a similar vein to RARR, \citet{bai2022constitutional} proposes to edit the outputs of LLMs using prompted LLMs to remove unsafe aspects of generated text.
As part of our future research, we aim to apply our denoising strategy to train efficient compact editors for addressing such undesired generation behaviors.

\paragraph{Distilling Large Language Models}
Given their generation prowess, LLMs have been incorporated into data generation pipelines, essentially distilling the knowledge of the language model if their outputs are used for training \citep{wang-etal-2021-want-reduce, bartolo-etal-2022-models, Lang2022CotrainingIP, Smith2022LanguageMI}.
\citet{Eisenstein2022HonestSF} follow a multi-step distillation pipeline like ours, chaining the outputs of multiple LLM calls and distilling the output into an explainable question answering model.
\citet{liu-etal-2022-wanli} uses the outputs of LLMs followed by filtering and human refinement to create WANLI, a challenging natural language inference dataset.
On the evaluation side, \citet{ribeiro-lundberg-2022-adaptive} use LLMs to generate evaluation sets for testing LLMs.
While similar, our denoising approach \textit{implicitly} distills the information in a large language model while simultaneously producing challenging training instances.
\section{Conclusion}\label{sec:conclusion}

Factuality and attribution are vital for the safe deployment of large language models.
However, these mechanisms are inherently lacking in LLMs.
Recent work has proposed augmenting the outputs of LLMs by retrieving evidence to attribute their outputs followed by prompting another LLM to edit the outputs to remove inconsistencies.
However, there is a heavy computational cost which bottleneck these methods which motivates a need to develop efficient editors, but this is hindered by training data scarcity.
To overcome these challenges, we use LLMs to corrupt text and fine-tune compact editors to denoise these faux hallucinations.
Our denoising method is completely unsupervised and our resulting editor, \approach{}, improves attribution performance across various datasets over prompted editors, while being order of magnitude faster to execute.

\bibliography{anthology,custom}

\begin{thebibliography}{46}
\expandafter\ifx\csname natexlab\endcsname\relax\def\natexlab#1{#1}\fi

\bibitem[{Aly et~al.(2021)Aly, Christodoulopoulos, Cocarascu, Guo, Mittal,
  Schlichtkrull, Thorne, and Vlachos}]{fever-2021-fact}
Rami Aly, Christos Christodoulopoulos, Oana Cocarascu, Zhijiang Guo, Arpit
  Mittal, Michael Schlichtkrull, James Thorne, and Andreas Vlachos, editors.
  2021.
\newblock \href {https://aclanthology.org/2021.fever-1.0} {\emph{Proceedings of
  the Fourth Workshop on Fact Extraction and VERification (FEVER)}}.
  Association for Computational Linguistics, Dominican Republic.

\bibitem[{Anantha et~al.(2021)Anantha, Vakulenko, Tu, Longpre, Pulman, and
  Chappidi}]{anantha-etal-2021-open}
Raviteja Anantha, Svitlana Vakulenko, Zhucheng Tu, Shayne Longpre, Stephen
  Pulman, and Srinivas Chappidi. 2021.
\newblock \href {https://doi.org/10.18653/v1/2021.naacl-main.44} {Open-domain
  question answering goes conversational via question rewriting}.
\newblock In \emph{Proceedings of the 2021 Conference of the North American
  Chapter of the Association for Computational Linguistics: Human Language
  Technologies}, pages 520--534, Online. Association for Computational
  Linguistics.

\bibitem[{Bai et~al.(2022)Bai, Kadavath, Kundu, Askell, Kernion, Jones, Chen,
  Goldie, Mirhoseini, McKinnon, Chen, Olsson, Olah, Hernandez, Drain, Ganguli,
  Li, Tran-Johnson, Perez, Kerr, Mueller, Ladish, Landau, Ndousse, Lukosuite,
  Lovitt, Sellitto, Elhage, Schiefer, Mercado, DasSarma, Lasenby, Larson,
  Ringer, Johnston, Kravec, Showk, Fort, Lanham, Telleen-Lawton, Conerly,
  Henighan, Hume, Bowman, Hatfield-Dodds, Mann, Amodei, Joseph, McCandlish,
  Brown, and Kaplan}]{bai2022constitutional}
Yuntao Bai, Saurav Kadavath, Sandipan Kundu, Amanda Askell, Jackson Kernion,
  Andy Jones, Anna Chen, Anna Goldie, Azalia Mirhoseini, Cameron McKinnon,
  Carol Chen, Catherine Olsson, Christopher Olah, Danny Hernandez, Dawn Drain,
  Deep Ganguli, Dustin Li, Eli Tran-Johnson, Ethan Perez, Jamie Kerr, Jared
  Mueller, Jeffrey Ladish, Joshua Landau, Kamal Ndousse, Kamile Lukosuite,
  Liane Lovitt, Michael Sellitto, Nelson Elhage, Nicholas Schiefer, Noemi
  Mercado, Nova DasSarma, Robert Lasenby, Robin Larson, Sam Ringer, Scott
  Johnston, Shauna Kravec, Sheer~El Showk, Stanislav Fort, Tamera Lanham,
  Timothy Telleen-Lawton, Tom Conerly, Tom Henighan, Tristan Hume, Samuel~R.
  Bowman, Zac Hatfield-Dodds, Ben Mann, Dario Amodei, Nicholas Joseph, Sam
  McCandlish, Tom Brown, and Jared Kaplan. 2022.
\newblock \href {http://arxiv.org/abs/2212.08073} {Constitutional ai:
  Harmlessness from ai feedback}.

\bibitem[{Balachandran et~al.(2022)Balachandran, Hajishirzi, Cohen, and
  Tsvetkov}]{Balachandran2022CorrectingDF}
Vidhisha Balachandran, Hannaneh Hajishirzi, William Cohen, and Yulia Tsvetkov.
  2022.
\newblock Correcting diverse factual errors in abstractive summarization via
  post-editing and language model infilling.
\newblock In \emph{Conference on Empirical Methods in Natural Language
  Processing}.

\bibitem[{Balachandran et~al.(2021)Balachandran, Vaswani, Tsvetkov, and
  Parmar}]{balachandran-etal-2021-simple}
Vidhisha Balachandran, Ashish Vaswani, Yulia Tsvetkov, and Niki Parmar. 2021.
\newblock \href {https://doi.org/10.18653/v1/2021.mrqa-1.16} {Simple and
  efficient ways to improve {REALM}}.
\newblock In \emph{Proceedings of the 3rd Workshop on Machine Reading for
  Question Answering}, pages 158--164, Punta Cana, Dominican Republic.
  Association for Computational Linguistics.

\bibitem[{Bartolo et~al.(2022)Bartolo, Thrush, Riedel, Stenetorp, Jia, and
  Kiela}]{bartolo-etal-2022-models}
Max Bartolo, Tristan Thrush, Sebastian Riedel, Pontus Stenetorp, Robin Jia, and
  Douwe Kiela. 2022.
\newblock \href {https://doi.org/10.18653/v1/2022.naacl-main.275} {Models in
  the loop: Aiding crowdworkers with generative annotation assistants}.
\newblock In \emph{Proceedings of the 2022 Conference of the North American
  Chapter of the Association for Computational Linguistics: Human Language
  Technologies}, pages 3754--3767, Seattle, United States. Association for
  Computational Linguistics.

\bibitem[{Bender et~al.(2021)Bender, Gebru, McMillan-Major, and
  Shmitchell}]{Bender2021OnTD}
Emily~M. Bender, Timnit Gebru, Angelina McMillan-Major, and Shmargaret
  Shmitchell. 2021.
\newblock On the dangers of stochastic parrots: Can language models be too big?
\newblock \emph{Proceedings of the 2021 ACM Conference on Fairness,
  Accountability, and Transparency}.

\bibitem[{Beyer et~al.(2021)Beyer, Zhai, Royer, Markeeva, Anil, and
  Kolesnikov}]{Beyer2021KnowledgeDA}
Lucas Beyer, Xiaohua Zhai, Am{\'e}lie Royer, Larisa Markeeva, Rohan Anil, and
  Alexander Kolesnikov. 2021.
\newblock Knowledge distillation: A good teacher is patient and consistent.
\newblock \emph{2022 IEEE/CVF Conference on Computer Vision and Pattern
  Recognition (CVPR)}, pages 10915--10924.

\bibitem[{Brown et~al.(2020)Brown, Mann, Ryder, Subbiah, Kaplan, Dhariwal,
  Neelakantan, Shyam, Sastry, Askell, Agarwal, Herbert-Voss, Krueger, Henighan,
  Child, Ramesh, Ziegler, Wu, Winter, Hesse, Chen, Sigler, Litwin, Gray, Chess,
  Clark, Berner, McCandlish, Radford, Sutskever, and
  Amodei}]{Brown2020LanguageMA}
Tom~B. Brown, Benjamin Mann, Nick Ryder, Melanie Subbiah, Jared Kaplan,
  Prafulla Dhariwal, Arvind Neelakantan, Pranav Shyam, Girish Sastry, Amanda
  Askell, Sandhini Agarwal, Ariel Herbert-Voss, Gretchen Krueger, T.~J.
  Henighan, Rewon Child, Aditya Ramesh, Daniel~M. Ziegler, Jeff Wu, Clemens
  Winter, Christopher Hesse, Mark Chen, Eric Sigler, Mateusz Litwin, Scott
  Gray, Benjamin Chess, Jack Clark, Christopher Berner, Sam McCandlish, Alec
  Radford, Ilya Sutskever, and Dario Amodei. 2020.
\newblock Language models are few-shot learners.
\newblock \emph{ArXiv}, abs/2005.14165.

\bibitem[{Cao et~al.(2020)Cao, Dong, Wu, and Cheung}]{cao-etal-2020-factual}
Meng Cao, Yue Dong, Jiapeng Wu, and Jackie Chi~Kit Cheung. 2020.
\newblock \href {https://doi.org/10.18653/v1/2020.emnlp-main.506} {Factual
  error correction for abstractive summarization models}.
\newblock In \emph{Proceedings of the 2020 Conference on Empirical Methods in
  Natural Language Processing (EMNLP)}, pages 6251--6258, Online. Association
  for Computational Linguistics.

\bibitem[{Carlini et~al.(2020)Carlini, Tram{\`e}r, Wallace, Jagielski,
  Herbert-Voss, Lee, Roberts, Brown, Song, Erlingsson, Oprea, and
  Raffel}]{Carlini2020ExtractingTD}
Nicholas Carlini, Florian Tram{\`e}r, Eric Wallace, Matthew Jagielski, Ariel
  Herbert-Voss, Katherine Lee, Adam Roberts, Tom~B. Brown, Dawn~Xiaodong Song,
  {\'U}lfar Erlingsson, Alina Oprea, and Colin Raffel. 2020.
\newblock Extracting training data from large language models.
\newblock In \emph{USENIX Security Symposium}.

\bibitem[{Chowdhery et~al.(2022)Chowdhery, Narang, Devlin, Bosma, Mishra,
  Roberts, Barham, Chung, Sutton, Gehrmann, Schuh, Shi, Tsvyashchenko, Maynez,
  Rao, Barnes, Tay, Shazeer, Prabhakaran, Reif, Du, Hutchinson, Pope, Bradbury,
  Austin, Isard, Gur-Ari, Yin, Duke, Levskaya, Ghemawat, Dev, Michalewski,
  Garc{\'i}a, Misra, Robinson, Fedus, Zhou, Ippolito, Luan, Lim, Zoph,
  Spiridonov, Sepassi, Dohan, Agrawal, Omernick, Dai, Pillai, Pellat,
  Lewkowycz, Moreira, Child, Polozov, Lee, Zhou, Wang, Saeta, D{\'i}az, Firat,
  Catasta, Wei, Meier-Hellstern, Eck, Dean, Petrov, and
  Fiedel}]{Chowdhery2022PaLMSL}
Aakanksha Chowdhery, Sharan Narang, Jacob Devlin, Maarten Bosma, Gaurav Mishra,
  Adam Roberts, Paul Barham, Hyung~Won Chung, Charles Sutton, Sebastian
  Gehrmann, Parker Schuh, Kensen Shi, Sasha Tsvyashchenko, Joshua Maynez,
  Abhishek Rao, Parker Barnes, Yi~Tay, Noam~M. Shazeer, Vinodkumar Prabhakaran,
  Emily Reif, Nan Du, Benton~C. Hutchinson, Reiner Pope, James Bradbury, Jacob
  Austin, Michael Isard, Guy Gur-Ari, Pengcheng Yin, Toju Duke, Anselm
  Levskaya, Sanjay Ghemawat, Sunipa Dev, Henryk Michalewski, Xavier Garc{\'i}a,
  Vedant Misra, Kevin Robinson, Liam Fedus, Denny Zhou, Daphne Ippolito, David
  Luan, Hyeontaek Lim, Barret Zoph, Alexander Spiridonov, Ryan Sepassi, David
  Dohan, Shivani Agrawal, Mark Omernick, Andrew~M. Dai,
  Thanumalayan~Sankaranarayana Pillai, Marie Pellat, Aitor Lewkowycz, Erica
  Moreira, Rewon Child, Oleksandr Polozov, Katherine Lee, Zongwei Zhou, Xuezhi
  Wang, Brennan Saeta, Mark D{\'i}az, Orhan Firat, Michele Catasta, Jason Wei,
  Kathleen~S. Meier-Hellstern, Douglas Eck, Jeff Dean, Slav Petrov, and Noah
  Fiedel. 2022.
\newblock Palm: Scaling language modeling with pathways.
\newblock \emph{ArXiv}, abs/2204.02311.

\bibitem[{Chung et~al.(2022)Chung, Hou, Longpre, Zoph, Tay, Fedus, Li, Wang,
  Dehghani, Brahma, Webson, Gu, Dai, Suzgun, Chen, Chowdhery, Valter, Narang,
  Mishra, Yu, Zhao, Huang, Dai, Yu, Petrov, hsin Chi, Dean, Devlin, Roberts,
  Zhou, Le, and Wei}]{Chung2022ScalingIL}
Hyung~Won Chung, Le~Hou, S.~Longpre, Barret Zoph, Yi~Tay, William Fedus, Eric
  Li, Xuezhi Wang, Mostafa Dehghani, Siddhartha Brahma, Albert Webson,
  Shixiang~Shane Gu, Zhuyun Dai, Mirac Suzgun, Xinyun Chen, Aakanksha
  Chowdhery, Dasha Valter, Sharan Narang, Gaurav Mishra, Adams~Wei Yu, Vincent
  Zhao, Yanping Huang, Andrew~M. Dai, Hongkun Yu, Slav Petrov, Ed~Huai hsin
  Chi, Jeff Dean, Jacob Devlin, Adam Roberts, Denny Zhou, Quoc Le, and Jason
  Wei. 2022.
\newblock Scaling instruction-finetuned language models.
\newblock \emph{ArXiv}, abs/2210.11416.

\bibitem[{Eisenstein et~al.(2022)Eisenstein, Andor, Bohnet, Collins, and
  Mimno}]{Eisenstein2022HonestSF}
Jacob Eisenstein, Daniel Andor, Bernd Bohnet, Michael Collins, and David Mimno.
  2022.
\newblock Honest students from untrusted teachers: Learning an interpretable
  question-answering pipeline from a pretrained language model.
\newblock \emph{ArXiv}, abs/2210.02498.

\bibitem[{Gao et~al.(2022)Gao, Dai, Pasupat, Chen, Chaganty, Fan, Zhao, Lao,
  Lee, Juan, and Guu}]{Gao2022RARRRA}
Luyu Gao, Zhuyun Dai, Panupong Pasupat, Anthony Chen, Arun~Tejasvi Chaganty,
  Yicheng Fan, Vincent Zhao, N.~Lao, Hongrae Lee, Da-Cheng Juan, and Kelvin
  Guu. 2022.
\newblock Rarr: Researching and revising what language models say, using
  language models.
\newblock \emph{ArXiv}, abs/2210.08726.

\bibitem[{Gehman et~al.(2020)Gehman, Gururangan, Sap, Choi, and
  Smith}]{gehman-etal-2020-realtoxicityprompts}
Samuel Gehman, Suchin Gururangan, Maarten Sap, Yejin Choi, and Noah~A. Smith.
  2020.
\newblock \href {https://doi.org/10.18653/v1/2020.findings-emnlp.301}
  {{R}eal{T}oxicity{P}rompts: Evaluating neural toxic degeneration in language
  models}.
\newblock In \emph{Findings of the Association for Computational Linguistics:
  EMNLP 2020}, pages 3356--3369, Online. Association for Computational
  Linguistics.

\bibitem[{Geva et~al.(2021)Geva, Khashabi, Segal, Khot, Roth, and
  Berant}]{geva-etal-2021-aristotle}
Mor Geva, Daniel Khashabi, Elad Segal, Tushar Khot, Dan Roth, and Jonathan
  Berant. 2021.
\newblock \href {https://doi.org/10.1162/tacl_a_00370} {Did aristotle use a
  laptop? a question answering benchmark with implicit reasoning strategies}.
\newblock \emph{Transactions of the Association for Computational Linguistics},
  9:346--361.

\bibitem[{He et~al.(2022)He, Zhang, and Roth}]{he_rethinking_2022}
Hangfeng He, Hongming Zhang, and Dan Roth. 2022.
\newblock \href {http://arxiv.org/abs/2301.00303} {Rethinking with {Retrieval}:
  {Faithful} {Large} {Language} {Model} {Inference}}.
\newblock \emph{ArXiv}.

\bibitem[{Iso et~al.(2020)Iso, Qiao, and Li}]{iso-etal-2020-fact}
Hayate Iso, Chao Qiao, and Hang Li. 2020.
\newblock \href {https://doi.org/10.18653/v1/2020.acl-main.17} {{F}act-based
  {T}ext {E}diting}.
\newblock In \emph{Proceedings of the 58th Annual Meeting of the Association
  for Computational Linguistics}, pages 171--182, Online. Association for
  Computational Linguistics.

\bibitem[{Iv et~al.(2022)Iv, Passos, Singh, and Chang}]{iv-etal-2022-fruit}
Robert Iv, Alexandre Passos, Sameer Singh, and Ming-Wei Chang. 2022.
\newblock \href {https://doi.org/10.18653/v1/2022.naacl-main.269} {{FRUIT}:
  Faithfully reflecting updated information in text}.
\newblock In \emph{Proceedings of the 2022 Conference of the North American
  Chapter of the Association for Computational Linguistics: Human Language
  Technologies}, pages 3670--3686, Seattle, United States. Association for
  Computational Linguistics.

\bibitem[{Izacard and Grave(2021)}]{izacard-grave-2021-leveraging}
Gautier Izacard and Edouard Grave. 2021.
\newblock \href {https://doi.org/10.18653/v1/2021.eacl-main.74} {Leveraging
  passage retrieval with generative models for open domain question answering}.
\newblock In \emph{Proceedings of the 16th Conference of the European Chapter
  of the Association for Computational Linguistics: Main Volume}, pages
  874--880, Online. Association for Computational Linguistics.

\bibitem[{Korbak et~al.(2023)Korbak, Shi, Chen, Bhalerao, Buckley, Phang,
  Bowman, and Perez}]{Korbak2023PretrainingLM}
Tomasz Korbak, Kejian Shi, Angelica Chen, Rasika Bhalerao, Christopher~L.
  Buckley, Jason Phang, Sam Bowman, and Ethan Perez. 2023.
\newblock Pretraining language models with human preferences.
\newblock \emph{ArXiv}, abs/2302.08582.

\bibitem[{Krishna et~al.(2021)Krishna, Roy, and
  Iyyer}]{krishna-etal-2021-hurdles}
Kalpesh Krishna, Aurko Roy, and Mohit Iyyer. 2021.
\newblock \href {https://doi.org/10.18653/v1/2021.naacl-main.393} {Hurdles to
  progress in long-form question answering}.
\newblock In \emph{Proceedings of the 2021 Conference of the North American
  Chapter of the Association for Computational Linguistics: Human Language
  Technologies}, pages 4940--4957, Online. Association for Computational
  Linguistics.

\bibitem[{Kwiatkowski et~al.(2019)Kwiatkowski, Palomaki, Redfield, Collins,
  Parikh, Alberti, Epstein, Polosukhin, Devlin, Lee, Toutanova, Jones, Kelcey,
  Chang, Dai, Uszkoreit, Le, and Petrov}]{kwiatkowski-etal-2019-natural}
Tom Kwiatkowski, Jennimaria Palomaki, Olivia Redfield, Michael Collins, Ankur
  Parikh, Chris Alberti, Danielle Epstein, Illia Polosukhin, Jacob Devlin,
  Kenton Lee, Kristina Toutanova, Llion Jones, Matthew Kelcey, Ming-Wei Chang,
  Andrew~M. Dai, Jakob Uszkoreit, Quoc Le, and Slav Petrov. 2019.
\newblock \href {https://doi.org/10.1162/tacl_a_00276} {Natural questions: A
  benchmark for question answering research}.
\newblock \emph{Transactions of the Association for Computational Linguistics},
  7:452--466.

\bibitem[{Lang et~al.(2022)Lang, Agrawal, Kim, and
  Sontag}]{Lang2022CotrainingIP}
Hunter Lang, Monica Agrawal, Yoon Kim, and David~A. Sontag. 2022.
\newblock Co-training improves prompt-based learning for large language models.
\newblock In \emph{International Conference on Machine Learning}.

\bibitem[{Liu et~al.(2022)Liu, Swayamdipta, Smith, and
  Choi}]{liu-etal-2022-wanli}
Alisa Liu, Swabha Swayamdipta, Noah~A. Smith, and Yejin Choi. 2022.
\newblock \href {https://aclanthology.org/2022.findings-emnlp.508} {{WANLI}:
  Worker and {AI} collaboration for natural language inference dataset
  creation}.
\newblock In \emph{Findings of the Association for Computational Linguistics:
  EMNLP 2022}, pages 6826--6847, Abu Dhabi, United Arab Emirates. Association
  for Computational Linguistics.

\bibitem[{Longpre et~al.(2021)Longpre, Perisetla, Chen, Ramesh, DuBois, and
  Singh}]{Longpre2021EntityBasedKC}
Shayne Longpre, Kartik~Kumar Perisetla, Anthony Chen, Nikhil Ramesh, Chris
  DuBois, and Sameer Singh. 2021.
\newblock Entity-based knowledge conflicts in question answering.
\newblock \emph{ArXiv}, abs/2109.05052.

\bibitem[{Maynez et~al.(2020)Maynez, Narayan, Bohnet, and
  McDonald}]{Maynez2020OnFA}
Joshua Maynez, Shashi Narayan, Bernd Bohnet, and Ryan~T. McDonald. 2020.
\newblock On faithfulness and factuality in abstractive summarization.
\newblock \emph{ArXiv}, abs/2005.00661.

\bibitem[{Min et~al.(2021)Min, Lewis, Zettlemoyer, and
  Hajishirzi}]{Min2021MetaICLLT}
Sewon Min, Mike Lewis, Luke Zettlemoyer, and Hannaneh Hajishirzi. 2021.
\newblock Metaicl: Learning to learn in context.
\newblock \emph{ArXiv}, abs/2110.15943.

\bibitem[{Raffel et~al.(2020)Raffel, Shazeer, Roberts, Lee, Narang, Matena,
  Zhou, Li, and Liu}]{raffel_exploring_2020}
Colin Raffel, Noam Shazeer, Adam Roberts, Katherine Lee, Sharan Narang, Michael
  Matena, Yanqi Zhou, Wei Li, and Peter~J. Liu. 2020.
\newblock \href {http://jmlr.org/papers/v21/20-074.html} {Exploring the
  {Limits} of {Transfer} {Learning} with a {Unified} {Text}-to-{Text}
  {Transformer}}.
\newblock \emph{Journal of Machine Learning Research}, 21(140):1--67.

\bibitem[{Rashkin et~al.(2021)Rashkin, Nikolaev, Lamm, Collins, Das, Petrov,
  Tomar, Turc, and Reitter}]{Rashkin2021MeasuringAI}
Hannah Rashkin, Vitaly Nikolaev, Matthew Lamm, Michael Collins, Dipanjan Das,
  Slav Petrov, Gaurav~Singh Tomar, Iulia Turc, and D.~Reitter. 2021.
\newblock Measuring attribution in natural language generation models.
\newblock \emph{ArXiv}, abs/2112.12870.

\bibitem[{Raunak et~al.(2021)Raunak, Menezes, and
  Junczys-Dowmunt}]{raunak-etal-2021-curious}
Vikas Raunak, Arul Menezes, and Marcin Junczys-Dowmunt. 2021.
\newblock \href {https://doi.org/10.18653/v1/2021.naacl-main.92} {The curious
  case of hallucinations in neural machine translation}.
\newblock In \emph{Proceedings of the 2021 Conference of the North American
  Chapter of the Association for Computational Linguistics: Human Language
  Technologies}, pages 1172--1183, Online. Association for Computational
  Linguistics.

\bibitem[{Ribeiro and Lundberg(2022)}]{ribeiro-lundberg-2022-adaptive}
Marco~Tulio Ribeiro and Scott Lundberg. 2022.
\newblock \href {https://doi.org/10.18653/v1/2022.acl-long.230} {Adaptive
  testing and debugging of {NLP} models}.
\newblock In \emph{Proceedings of the 60th Annual Meeting of the Association
  for Computational Linguistics (Volume 1: Long Papers)}, pages 3253--3267,
  Dublin, Ireland. Association for Computational Linguistics.

\bibitem[{Sanh et~al.(2021)Sanh, Webson, Raffel, Bach, Sutawika, Alyafeai,
  Chaffin, Stiegler, Scao, Raja, Dey, Bari, Xu, Thakker, Sharma, Szczechla,
  Kim, Chhablani, Nayak, Datta, Chang, Jiang, Wang, Manica, Shen, Yong, Pandey,
  Bawden, Wang, Neeraj, Rozen, Sharma, Santilli, F{\'e}vry, Fries, Teehan,
  Biderman, Gao, Bers, Wolf, and Rush}]{Sanh2021MultitaskPT}
Victor Sanh, Albert Webson, Colin Raffel, Stephen~H. Bach, Lintang Sutawika,
  Zaid Alyafeai, Antoine Chaffin, Arnaud Stiegler, Teven~Le Scao, Arun Raja,
  Manan Dey, M~Saiful Bari, Canwen Xu, Urmish Thakker, Shanya Sharma, Eliza
  Szczechla, Taewoon Kim, Gunjan Chhablani, Nihal~V. Nayak, Debajyoti Datta,
  Jonathan Chang, Mike Tian-Jian Jiang, Han Wang, Matteo Manica, Sheng Shen,
  Zheng~Xin Yong, Harshit Pandey, Rachel Bawden, Thomas Wang, Trishala Neeraj,
  Jos Rozen, Abheesht Sharma, Andrea Santilli, Thibault F{\'e}vry, Jason~Alan
  Fries, Ryan Teehan, Stella~Rose Biderman, Leo Gao, Tali Bers, Thomas Wolf,
  and Alexander~M. Rush. 2021.
\newblock Multitask prompted training enables zero-shot task generalization.
\newblock \emph{ArXiv}, abs/2110.08207.

\bibitem[{Schick et~al.(2022)Schick, Dwivedi-Yu, Jiang, Petroni, Lewis,
  Izacard, You, Nalmpantis, Grave, and Riedel}]{Schick2022PEERAC}
Timo Schick, Jane Dwivedi-Yu, Zhengbao Jiang, Fabio Petroni, Patrick Lewis,
  Gautier Izacard, Qingfei You, Christoforos Nalmpantis, Edouard Grave, and
  Sebastian Riedel. 2022.
\newblock Peer: A collaborative language model.
\newblock \emph{ArXiv}, abs/2208.11663.

\bibitem[{Smith et~al.(2022)Smith, Fries, Hancock, and
  Bach}]{Smith2022LanguageMI}
Ryan Smith, Jason~Alan Fries, Braden Hancock, and Stephen~H. Bach. 2022.
\newblock Language models in the loop: Incorporating prompting into weak
  supervision.
\newblock \emph{ArXiv}, abs/2205.02318.

\bibitem[{Stanton et~al.(2021)Stanton, Izmailov, Kirichenko, Alemi, and
  Wilson}]{Stanton2021DoesKD}
Samuel Stanton, Pavel Izmailov, P.~Kirichenko, Alexander~A. Alemi, and
  Andrew~Gordon Wilson. 2021.
\newblock Does knowledge distillation really work?
\newblock \emph{ArXiv}, abs/2106.05945.

\bibitem[{Stiennon et~al.(2020)Stiennon, Ouyang, Wu, Ziegler, Lowe, Voss,
  Radford, Amodei, and Christiano}]{Stiennon2020LearningTS}
Nisan Stiennon, Long Ouyang, Jeff Wu, Daniel~M. Ziegler, Ryan~J. Lowe, Chelsea
  Voss, Alec Radford, Dario Amodei, and Paul Christiano. 2020.
\newblock Learning to summarize from human feedback.
\newblock \emph{ArXiv}, abs/2009.01325.

\bibitem[{Thoppilan et~al.(2022)Thoppilan, Freitas, Hall, Shazeer,
  Kulshreshtha, Cheng, Jin, Bos, Baker, Du, Li, Lee, Zheng, Ghafouri, Menegali,
  Huang, Krikun, Lepikhin, Qin, Chen, Xu, Chen, Roberts, Bosma, Zhou, Chang,
  Krivokon, Rusch, Pickett, Meier-Hellstern, Morris, Doshi, Santos, Duke,
  S{\o}raker, Zevenbergen, Prabhakaran, D{\'i}az, Hutchinson, Olson, Molina,
  Hoffman-John, Lee, Aroyo, Rajakumar, Butryna, Lamm, Kuzmina, Fenton, Cohen,
  Bernstein, Kurzweil, Aguera-Arcas, Cui, Croak, hsin Chi, and
  Le}]{Thoppilan2022LaMDALM}
Romal Thoppilan, Daniel~De Freitas, Jamie Hall, Noam~M. Shazeer, Apoorv
  Kulshreshtha, Heng-Tze Cheng, Alicia Jin, Taylor Bos, Leslie Baker, Yu~Du,
  Yaguang Li, Hongrae Lee, Huaixiu Zheng, Amin Ghafouri, Marcelo Menegali,
  Yanping Huang, Maxim Krikun, Dmitry Lepikhin, James Qin, Dehao Chen,
  Yuanzhong Xu, Zhifeng Chen, Adam Roberts, Maarten Bosma, Yanqi Zhou,
  Chung-Ching Chang, I.~A. Krivokon, Willard~James Rusch, Marc Pickett,
  Kathleen~S. Meier-Hellstern, Meredith~Ringel Morris, Tulsee Doshi,
  Renelito~Delos Santos, Toju Duke, Johnny~Hartz S{\o}raker, Ben Zevenbergen,
  Vinodkumar Prabhakaran, Mark D{\'i}az, Ben Hutchinson, Kristen Olson,
  Alejandra Molina, Erin Hoffman-John, Josh Lee, Lora Aroyo, Ravindran
  Rajakumar, Alena Butryna, Matthew Lamm, V.~O. Kuzmina, Joseph Fenton, Aaron
  Cohen, Rachel Bernstein, Ray Kurzweil, Blaise Aguera-Arcas, Claire Cui,
  Marian Croak, Ed~Huai hsin Chi, and Quoc Le. 2022.
\newblock Lamda: Language models for dialog applications.
\newblock \emph{ArXiv}, abs/2201.08239.

\bibitem[{Thorne and Vlachos(2020)}]{Thorne2020EvidencebasedFE}
James Thorne and Andreas Vlachos. 2020.
\newblock Evidence-based factual error correction.
\newblock In \emph{Annual Meeting of the Association for Computational
  Linguistics}.

\bibitem[{Touvron et~al.(2023)Touvron, Lavril, Izacard, Martinet, Lachaux,
  Lacroix, Rozi{\`e}re, Goyal, Hambro, Azhar, Rodriguez, Joulin, Grave, and
  Lample}]{Touvron2023LLaMAOA}
Hugo Touvron, Thibaut Lavril, Gautier Izacard, Xavier Martinet, Marie-Anne
  Lachaux, Timoth{\'e}e Lacroix, Baptiste Rozi{\`e}re, Naman Goyal, Eric
  Hambro, Faisal Azhar, Aur'elien Rodriguez, Armand Joulin, Edouard Grave, and
  Guillaume Lample. 2023.
\newblock Llama: Open and efficient foundation language models.
\newblock \emph{ArXiv}, abs/2302.13971.

\bibitem[{Wang et~al.(2021)Wang, Liu, Xu, Zhu, and
  Zeng}]{wang-etal-2021-want-reduce}
Shuohang Wang, Yang Liu, Yichong Xu, Chenguang Zhu, and Michael Zeng. 2021.
\newblock \href {https://doi.org/10.18653/v1/2021.findings-emnlp.354} {Want to
  reduce labeling cost? {GPT}-3 can help}.
\newblock In \emph{Findings of the Association for Computational Linguistics:
  EMNLP 2021}, pages 4195--4205, Punta Cana, Dominican Republic. Association
  for Computational Linguistics.

\bibitem[{Wei et~al.(2022)Wei, Wang, Schuurmans, Bosma, ichter, Xia, Chi, Le,
  and Zhou}]{wei_chain--thought_2022}
Jason Wei, Xuezhi Wang, Dale Schuurmans, Maarten Bosma, brian ichter, Fei Xia,
  Ed~Chi, Quoc~V Le, and Denny Zhou. 2022.
\newblock \href
  {https://proceedings.neurips.cc/paper_files/paper/2022/file/9d5609613524ecf4f15af0f7b31abca4-Paper-Conference.pdf}
  {Chain-of-{Thought} {Prompting} {Elicits} {Reasoning} in {Large} {Language}
  {Models}}.
\newblock In \emph{Advances in {Neural} {Information} {Processing} {Systems}},
  volume~35, pages 24824--24837. Curran Associates, Inc.

\bibitem[{Ye et~al.(2022)Ye, Kim, Jang, Shin, and Seo}]{Ye2022GuessTI}
Seonghyeon Ye, Doyoung Kim, Joel Jang, Joongbo Shin, and Minjoon Seo. 2022.
\newblock Guess the instruction! flipped learning makes language models
  stronger zero-shot learners.
\newblock \emph{ArXiv}, abs/2210.02969.

\bibitem[{Zhao et~al.(2021)Zhao, Wallace, Feng, Klein, and
  Singh}]{Zhao2021CalibrateBU}
Tony Zhao, Eric Wallace, Shi Feng, Dan Klein, and Sameer Singh. 2021.
\newblock Calibrate before use: Improving few-shot performance of language
  models.
\newblock \emph{ArXiv}, abs/2102.09690.

\bibitem[{Ziegler et~al.(2019)Ziegler, Stiennon, Wu, Brown, Radford, Amodei,
  Christiano, and Irving}]{Ziegler2019FineTuningLM}
Daniel~M. Ziegler, Nisan Stiennon, Jeff Wu, Tom~B. Brown, Alec Radford, Dario
  Amodei, Paul Christiano, and Geoffrey Irving. 2019.
\newblock Fine-tuning language models from human preferences.
\newblock \emph{ArXiv}, abs/1909.08593.

\end{thebibliography}
\appendix\label{sec:appendix}

\section{Prompts for Creating Training Data}\label{sec:appendix:prompts}

{ %

\lstdefinelanguage{prompt}{
comment=[s][\textbf]{\{}{\}},   %
}
\lstset{
language=prompt,
basicstyle=\tiny\ttfamily,
breaklines=true,
breakindent=0pt,
numbers=left,
numberstyle=\tiny\ttfamily\color{gray},
frame=single,
numbersep=10pt,
xleftmargin=20pt,
framexleftmargin=20pt,
}

\begin{figure*}[p]
\begin{center}
\begin{lstlisting}
Summarize all the pieces of text. Paraphrase the text and change the syntax.

{text}

Summary:
\end{lstlisting}
\caption{
    Zero-shot prompt for multi-document summarization.
    The input \texttt{\{text\}} can be multiple pieces of text from different sources separated by a new-line.
}
\label{fig:prompt_summarize}
\end{center}
\end{figure*}

\begin{figure*}[p]
\begin{center}
\begin{lstlisting}
Corrupt the text by first generating a reasoning that describes what you will change, then following the reasoning to change the the text. Make the reasoning false but believable. Do not remove any information.

Text: The new revelation came Monday as the Department of Justice filed federal charges of assault and attempted kidnapping against the man suspected of attacking Paul Pelosi.
Number of things to change: 1.
Reasoning: I am going to swap "the man" and "Paul Pelosi".
Corruption: The new revelation came Monday as Paul Pelosi filed federal charges of assault and attempted kidnapping against Paul Pelosi suspected of attacking the man.

Text: Grapes of Wrath is a novel published in 1939, written by John Steinbeck. It takes place during the Great Depression, and focuses on the Joad family and their journey from Oklahoma to California.
Number of things to change: 3.
Reasoning: I am going to change when the novel was published to "1937", what it focuses on to "class discrimination", and add the fact that John Steinbeck was British.
Corruption: Grapes of Wrath is a novel published in 1937, written by the British author John Steinbeck. It takes place during the Great Depression, and focuses on the class discrimination on display.

Text: {text}
Number of things to change: {num_corruptions}.
Reasoning:
\end{lstlisting}
\caption{
    Few-shot prompt for corruption.
    Our corruption happens in a chain-of-thought fashion \citep{wei_chain--thought_2022}, allowing more flexibility in determining how many pieces of information to corrupt and what kinds of information to change.
}
\label{fig:prompt_corruption}
\end{center}
\end{figure*}

}
\end{document}